\documentclass[conference]{IEEEtran}
\IEEEoverridecommandlockouts

\usepackage{cite}
\usepackage{amsmath,amssymb,amsfonts}
\usepackage{algorithmic}
\usepackage{graphicx}
\usepackage{textcomp}

\usepackage{booktabs}
\usepackage{hyperref}
\hypersetup{hypertex=true,
colorlinks=true,
linkcolor=blue,
anchorcolor=blue,
citecolor=blue}
\usepackage{multirow}
\usepackage{stfloats} 

\def\BibTeX{{\rm B\kern-.05em{\sc i\kern-.025em b}\kern-.08em
    T\kern-.1667em\lower.7ex\hbox{E}\kern-.125emX}}
\makeatletter
\newcommand{\linebreakand}{%
  \end{@IEEEauthorhalign}
  \hfill\mbox{}\par
  \mbox{}\hfill\begin{@IEEEauthorhalign}
}
\makeatother
\DeclareRobustCommand*{\IEEEauthorrefmark}[1]{%
    \raisebox{0pt}[0pt][0pt]{\textsuperscript{\footnotesize\ensuremath{#1}}}}

\begin{document}

\title{Are Large Language Models Table-based Fact-Checkers?}

\author{
\IEEEauthorblockN{
Hanwen Zhang\IEEEauthorrefmark{1,2},
Qingyi Si\IEEEauthorrefmark{1,2}, 
Peng Fu\thanks{* Corresponding author.}\IEEEauthorrefmark{1,2*},
Zheng Lin\IEEEauthorrefmark{1,2},and
Weiping Wang\IEEEauthorrefmark{1}}

\IEEEauthorblockA{\IEEEauthorrefmark{1}Institute of Information Engineering, Chinese Academy of Sciences, Beijing, China}
\IEEEauthorblockA{\IEEEauthorrefmark{2}School of Cyber Security, University of Chinese Academy of Sciences, Beijing, China}
\IEEEauthorblockA{\{zhanghanwen, siqingyi, fupeng, linzheng, wangweiping\}@iie.ac.cn}
}

\maketitle

\begin{abstract}
Table-based Fact Verification (TFV) aims to extract the entailment relationship between statements and structured tables. Existing TFV methods based on small-scale models suffer from insufficient labeled data and weak zero-shot ability. Recently, the appearance of Large Language Models (LLMs) has gained lots of attraction in research fields. They have shown strong zero-shot and in-context learning capabilities on several NLP tasks, but their potential on TFV is still unknown. In this work, we implement a preliminary study on whether LLMs are table-based fact-checkers. In detail, we design various prompts to explore how in-context learning can help LLMs in TFV, i.e., zero-shot and few-shot TFV capability. 
Besides, we carefully design and construct TFV instructions to investigate the performance gain brought by the instruction tuning of LLMs. 
Experimental results demonstrate that LLMs can achieve acceptable results on zero-shot and few-shot TFV with prompt engineering, while instruction-tuning can stimulate the TFV capability significantly. We also make some valuable findings about the format of zero-shot prompts and the number of in-context examples. Finally, we analyze some possible directions to promote the accuracy of TFV via LLMs, which will benefit further research on table reasoning. 
\end{abstract}


\section{Introduction}
\begin{figure}[]
\centerline{\includegraphics[width=0.7\linewidth]{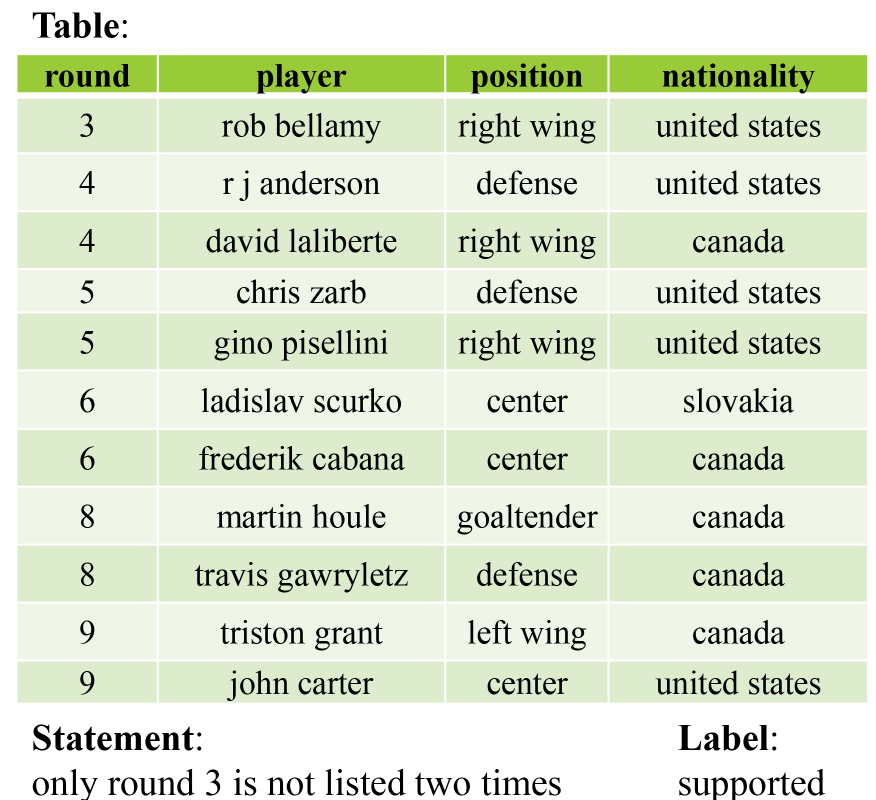}}
\caption{An example of the Table-based Fact Verification task.}
\label{fig:1-1}
\end{figure}

Fact verification aiming to determine the veracity of given claims has been increasingly popular for downstream applications like fake news detection and misinformation identification. Initial studies on fact verification focused on unstructured textual evidence \cite{fever}, while evidence with table format has attracted more attention in recent years\cite{tabfact,infotabs}. Table-based fact verification (TFV) intends to identify whether a statement is supported or refuted by the structured or semi-structured table, which is more challenging due to the rich structure information of tables and complex reasoning types of statements. Fig.\ref{fig:1-1} shows an example of table-based fact verification. To verify the statement ``only round 3 is not listed two times'', we should first locate the ``round" column in the table, and then count the number of occurrences for each element.

Previous studies have mainly proposed two kinds of methods for table-based fact verification: (1) program-based methods\cite{LFC,progvgat,lergv,tabfact} and (2) pretrain-based methods\cite{tapas,tapex,tableformer,samoe,pasta}. Program-based methods utilize the semantic parser trained under weakly-supervised settings to transform statements into logical forms with logical and numerical semantics. The pretrain-based methods retrain the text-based pre-trained language models (PLMs) on table-aware pre-training tasks\cite{tapas,tapex,pasta} or extra model structure\cite{samoe} to enhance the reasoning ability on tables. However, both two kinds of methods rely on sufficient training data and have difficulties explaining their judgment.  

Meanwhile, the closed-source large language models (LLMs) such as GPT-3\cite{gpt3}, Codex\cite{codex}, and ChatGPT demonstrate a marvelous 
in-context learning ability (especially zero-shot and few-shot learning)
to achieve competitive performances in various NLP tasks\cite{stance,summarization,translation}. But they can be used only by calling API interfaces with designed prompts. The open-source of LLaMA\cite{llama}, a competitive LLM, and the success in instruction tuning of LLMs\cite{alpaca,instructgpt} make it possible to deploy and customize LLM for various tasks. 
The closed-source and open-source LLMs can respectively promote research directions of zero-/few-shot learning and instruction tuning for table-based face verification. This area of research is still in its infancy.
Existing related works\cite{binder,datar} only regard LLMs as an external knowledge base, which is utilized to improve small-scale models. To date, no research has been conducted on the direct use of LLMs for completing TFV tasks. This leaves the question  ``Are large language models table-based fact-checkers" unanswered. 

To answer the question above, this paper conducts comprehensive research on LLMs for table-based fact verification under three settings. First, we explore the performance of LLMs under the zero-shot setting, which represents the inherent TFV capability of LLMs. Second, we use LLMs to achieve TFV in few-shot fashion, which reflects how in-context learning promotes TFV. Third, we construct the TFV instructions for tuning LLaMA to induce the full potential of LLMs in TFV. 

We conduct extensive experiments on the above three settings, leading to valuable findings and conclusions. For zero-shot learning of LLMs in TFV, we discover a more robust design of prompts. For few-shot learning, we find that more in-context examples do not necessarily lead to better results, and 2-shot is a good choice. However, the prompt engineering under the above two in-context learning settings cannot significantly improve LLMs' TFV capability, while instruction-tuning of LLMs can. Finally, we discuss some potential research directions for improving the performance of TFV via LLMs in future studies. The result files can be accessed at the following URL: \url{https://github.com/Heaven-zhw/LLM-on-tabfact}.

\section{Related work}
\noindent\textbf{Table-based Fact Verification.}
There are two existing mainstream methods for TFV: program-based and pretrain-based. Program-based methods either synthesize latent programs to access structured tables\cite{tabfact} or utilize programs indirectly to get the logical and numerical semantics\cite{LFC,progvgat}. 
Pretrain-based methods, inspired by the success of general PLMs in natural language tasks, are proposed to learn various table tasks jointly \cite{tapas,tapex,pasta} for better table understanding. These models are pre-trained with newly designed table-specific tasks (e.g., SQL execution\cite{tapex} and sentence-table cloze\cite{pasta}) on a massive table and natural language data \cite{tapas,pasta}.   
More recently, some works \cite{binder,datar} also leverage LLMs in their methods. However, they serve LLMs as auxiliary components (e.g., semantic parser\cite{binder}, evidence decomposers\cite{datar}) to improve the small-scale models. The direct use of LLMs as table-based fact-checkers has not yet been explored.  

\noindent\textbf{Prompt Engineering of Large Language Models.}
Prompt engineering is an important topic in the era of LLMs. 
Since the impressive few-shot capabilities demonstrated by GPT3 \cite{gpt3}, researchers have been continuously attempting to guide various abilities of LLMs through prompt design, also known as in-context learning. The emergence of ChatGPT further promotes the development of prompt engineering. Because these competitive LLMs have been closed source for a long time (until the leak of LLaMA), researchers can only achieve specific tasks \cite{stance, summarization, translation} by calling API interfaces to transmit designed prompts. This paper attempts to leverage LLMs in the TFV task under zero-shot and few-shot settings. 

\noindent\textbf{Instruction Tuning of Large Language Models.}
Base LLMs are primarily pre-trained to predict the next word, which can be applied to massive unsupervised texts. This pre-training stage is the main source of accumulating knowledge to understand and generate language. Recently, the open-source community \cite{alpaca,si2023empirical} has found that instruction fine-tuning base LLMs with specific instructions can significantly improve their ability to follow instructions and better accomplish complex tasks. 
However, fine-tuning language models (LLMs) for specific tasks often requires huge computing power. Parameter-efficient fine-tuning (PEFT) methods\cite{lora,ptuning} make it possible to fine-tune LLMs at a lower cost. We followed these widely used strategies to fine-tune LLMs for the TFV task.

\section{Large Language Models for TFV}
\subsection{Task Definition}

Here is a symbolized definition of the table-based fact verification task: Given a structured table $T=\{t_{ij}|i\le m, j\le n\}$ with $m$ rows and $n$ columns and a statement $S=\{w_i|i\le W_S\}$ with $W_S$ words, we need a verification model $f$ to predict a verdict label $\hat{y} = f(S, T)$,  usually $\hat{y}\in \{0,1\}$ for binary classification datasets. 

As the blossom of LLMs for several tasks, we wonder whether LLMs are good table-based fact-checkers. To answer this question, we induce LLMs to directly generate the final answer for TFV. We have explored their capability in TFV under the in-context learning (i.e, zero-shot and few-shot prompt engineering) and instruction tuning settings, which will be discussed in Section \ref{sec4}, Section \ref{sec5}, and Section \ref{sec6} individually. 


\subsection{Dataset and Experimental Settings} 

We evaluate LLMs on TabFact\cite{tabfact}, a TFV benchmark dataset, including 16K tables from Wikipedia for 118K natural language statements written by crowdworkers. The statements are labeled from the collection \{supported, refuted\}. The whole dataset is randomly divided into train, development, and test subsets with the ratio of 8:1:1, wherein a small test set with 1998 samples is further divided from the original test set for human evaluation. To facilitate comparison with humans and reduce the cost, we use the small test for our experiments, noted as ``TabFact-small-test". 

For evaluation, we use accuracy as the evaluation metric followed by the original setting. We extract the predicted label from the generative content by keyword matching, but sometimes we can't extract the keywords from the responses. In all our experiments, we regard these unrelated or empty responses as REFUTED class.


This paper mainly uses representative LLMs, ChatGPT, and LLaMA to conduct experiments. The former is the most popular LLM base in the zero-/few-shot learning research of LLMs, which has sparked a series of ``API" research in prompt engineering. The latter is the most competitive open-source LLM, which has important significance in instruction-tuning research.

For ChatGPT, we conduct the experiments by API version of GPT-3.5-turbo. For LLaMA, the original version\cite{llama}, the iterative and length-enhanced version\cite{llama2}, and a further fine-tuned and dialogue-adaptive version are involved in our experiments, namely noted as ``LLaMA-1'', ``LLaMA-2'' and ``LLaMA-2-chat'', the latter two of which belong to LLaMA2 family. All versions of LLaMA are 7B parameters. According to the length of context windows, we set the max length of input to 2048 for LLaMA-1, 4096 for LLaMA-2, and LLaMA-2-chat.

\section{In-context Learning of LLMs} 

In this section, we adopt the prompt engineering strategies to observe the gain brought by the in-context learning ability of LLMs on table-based fact verification. 
\subsection{Zero-shot Learning of LLMs} \label{sec4}
To leverage these generative models, we need to design elaborate prompts for TFV to stimulate the verification ability of LLMs. Under the zero-shot setting, there is nothing but the TFV task description to provide for LLMs. We explore the following zero-shot prompts for TFV.

\noindent\textbf{1. Sentence}. Prompts with one-turn dialogue are the simplest ways to implement ChatGPT on NLP tasks. It just needs to input one sequence and collect the response as the result. In terms of TFV, we can simply construct a long sentence with the task description, the linearized table, and the statement to input into ChatGPT, forming the \textbf{Sentence} prompt. The form of the Sentence prompt is ``\textit{Give you a statement and a table, please tell me whether the statement is supported or refuted by the table. The table is [LINEARIZED TABLE]. The statement is [STATEMENT].}".

\noindent\textbf{2. Sentence+word}.  Considering zero-shot LLMs may generate responses with uncertain length or undesired label words, we can input an extra description to constrain the format of the output. Inspired by \cite{summarization}, we append an additional word-guided sentence ``\textit{Just answer only one word ``supported" or ``refuted" without other tokens.}" after Sentence prompt, named \textbf{Sentence+word}. With word-guided sentences, it's hoped to extract the verdict label more conveniently. 
 
\begin{figure}[h]
\centering

\includegraphics[width=1.0\linewidth]{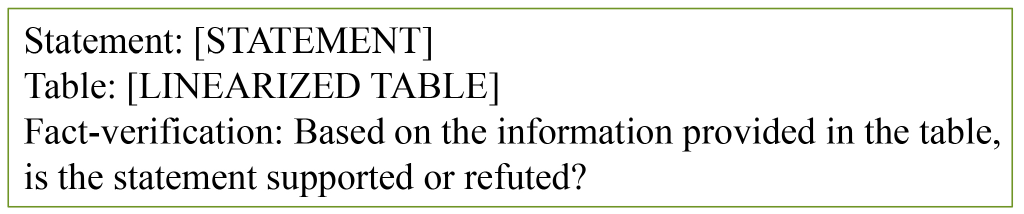}
\caption{Paragraph prompt}
\label{fig:4-1}
\end{figure}

\noindent\textbf{3. Paragraph}. Additionally, inspired by \cite{translator}, we ask the web version of ChatGPT(ChatGPT-web) to provide TFV prompts. From the responses, we choose an accessible and related prompt named \textbf{Paragraph} prompt as it consists of several paragraphs. The Paragraph prompt is shown as Fig.\ref{fig:4-1}.

\begin{figure}[t]
\centering

\includegraphics[width=1.0\linewidth]{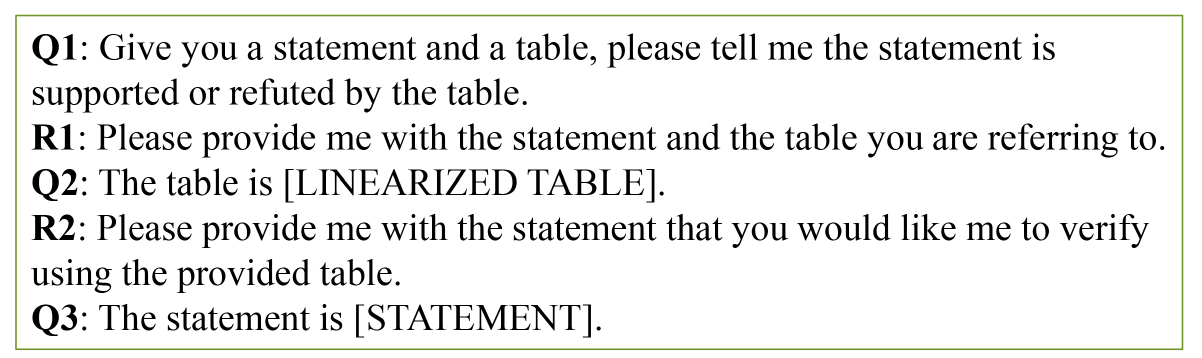}
\caption{Dialogue-simple prompt}
\label{fig:4-2}
\end{figure}

\noindent\textbf{4. Dialogue}. A more interactive way to use dialogue models is multi-turn dialogue, which inputs all dialogue history and extracts results from the response of the last turn. For TFV, we can exploit a multi-turn dialogue to input task description, table, and statement separately. To simulate real responses, we input requests into ChatGPT-web and extract some responses. Here we will use two kinds of responses generated by ChatGPT-web to fill the prompt, named \textbf{Dialogue-simple} and \textbf{Dialogue-complex}. The Dialogue-simple prompt is demonstrated as Fig.\ref{fig:4-2}. For the Dialogue-complex prompt, the questions are the same as the Dialogue-simple, while the responses R1 and R2 are replaced with more courteous and indicative sentences. In the Dialogue-complex prompt, R1 is ``\textit{I'll do my best to help you with that! Please provide me with the statement and the table, and I'll let you know if the statement is supported or refuted by the information presented in the table.}", and R2 is ``\textit{Thank you for providing the table. Please provide me with the statement that you want me to check against the table, and I'll let you know if it's supported or refuted by the information in the table.}".

\begin{figure}[t]
\centering

\includegraphics[width=1.0\linewidth]{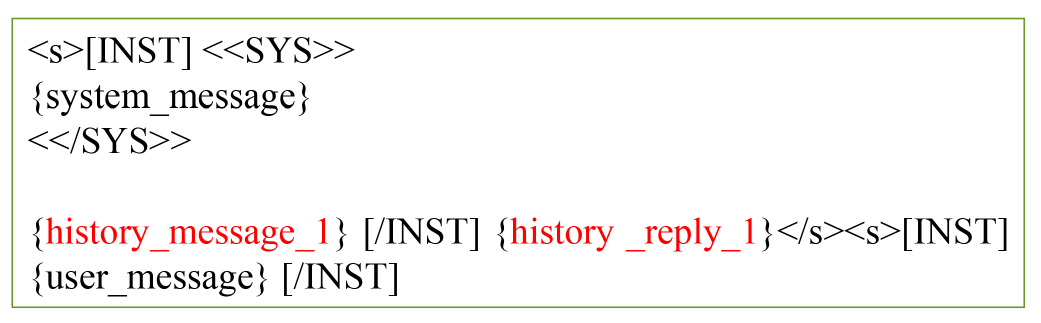}
\caption{Recommended Instruction prompt of LLaMA-2}
\label{fig:4-3}
\end{figure}

It is worth noticing how to input the dialogue history to LLMs. For ChatGPT, it's convenient to add dialogue history to API by switching the role to ``user" for history input or ``assistant" for history output. For LLaMA, we use the recommended dialogue instruction prompt \footnote{https://huggingface.co/blog/llama2\#how-to-prompt-llama-2} (as shown in Figure \ref{fig:4-3}) to imitate a dialogue environment, which is also employed in the fine-tuning procedure of LLaMA-2-chat, a dialogue adaptive version of LLaMA-2. Therefore, we select LLaMA-2-chat and ChatGPT to perform the zero-shot TFV experiment. The accuracy of different methods under zero-shot setting on TabFact-small-test is shown in Table \ref{zeroshot-result}.

\begin{table}[h]
\centering
\caption{Zero-shot Experimental results}
\begin{tabular}{llc}
   \toprule
   Model & Prompt & Accuracy(\%) \\
   \midrule
   \multirow{5}{*}{LLaMA-2-chat} & Sentence & 53.1 \\
   & Sentence+word & 50.0 \\
   & Paragraph & 55.5 \\
   & Dialogue-simple & 51.5 \\
   & Dialogue-complex & 55.6 \\
   \midrule
   \multirow{5}{*}{ChatGPT}  & Sentence & 73.6 \\
   & Sentence+word & 69.7 \\
   & Paragraph & 57.8 \\
   & Dialogue-simple & 74.2 \\
   & Dialogue-complex & 74.9 \\
   \midrule
   Random Guessing & - & 50.0 \\
   Human & - & 92.1 \\
   \bottomrule
\end{tabular}
\label{zeroshot-result}
\end{table}

The results show that ChatGPT demonstrates an acceptable performance under the zero-shot setting, which is comparable to early baseline methods. However, the accuracy of LLaMA-2-chat with all prompts is slightly over 0.5, the result of random guessing for binary classification tasks. We analyze that compared with LLaMA (7B), ChatGPT has more parameters and has seen more tabular data within pre-training. Zero-shot LLaMA-2-chat can generate seemingly reasonable responses, but it lacks reasoning ability on tables.

For concrete prompts, there are some interesting findings as follows. First, the Dialogue-complex prompt shows a superior performance among these prompts, which can be attributed to the greater number of interactions they facilitate. Second, the Sentence+word prompt under-performs the Sentence prompt with both LLaMA and ChatGPT. This is mainly because word-guided prompt discourages LLMs from thinking deeply. Third, we notice the Paragraph prompt with ChatGPT reaches an unexpectedly low accuracy. One possible reason is that the Paragraph prompt deviates from a smooth and natural expression of dialogue, thereby exerting a greater influence on ChatGPT for zero-shot learning.

\subsection{Few-shot Learning of LLMs} \label{sec5}

Besides task description, a few in-context TFV examples including inputs and labels are provided under the few-shot setting. Different from previous few-shot learning methods for small models, LLMs can imitate the response pattern from a small amount of in-context examples without updating model parameters. Here we assess the in-context learning ability of LLMs on TFV.

\begin{figure}[h]
\centering
\includegraphics[width=1.0\linewidth]{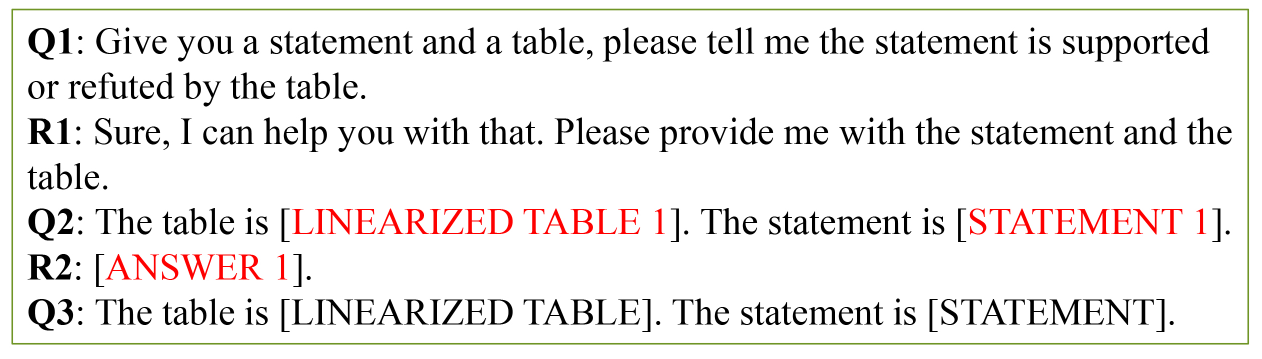}
\caption{Dialogue-style in-context prompt (1-shot)}
\label{fig:5-1}
\end{figure}

We adapt a dialogue-style in-context prompt for few-shot experiments as Fig.\ref{fig:5-1} shows. For the N-shot condition, the number of dialogue turns can be expanded by introducing more question-response pairs after R1, followed by the example that needs to be predicted at the end. 

We choose LLaMA-2, LLaMA-2-chat, and ChatGPT for our few-shot experiments. Same as zero-shot experiments, LLaMA-dialogue instruction prompts (Fig.\ref{fig:4-3}) are used for LLaMA-2 and LLaMA-2-chat. Besides, in-context examples are handpicked from the training set and remain consistent across various LLMs with the same shot number. We opt for the best choice of in-context examples among 4 selected candidates under the 1-shot and 2-shot settings.

Table \ref{fewshot-result} shows the accuracy of different methods under the few-shot setting on TabFact-small-test. Compared with the zero-shot setting, ChatGPT gains better accuracy with the assistance of in-context examples. The victory over Codex demonstrates the superior in-context learning capability of ChatGPT. 

However, the LLaMA models still perform poorly with few in-context samples. The accuracy attained by LLaMA-2-chat hovers slightly above 0.5, showing no significant improvement compared with the zero-shot performance. In some cases, LLaMA-2 even performs worse than random guessing. This indicates that LLaMA models with 7B parameters are inadequate in learning sufficient reasoning ability from TFV in-context examples. 

Additionally, we observe a decline in accuracy as the number of in-context examples increases from 2 to 4. A possible reason is the excessive length of the in-context examples. Due to the limited contextual memory capability of ChatGPT and LLaMA, an excess of inputs may bring a burden to inference.

\begin{table}[h]
\centering
\caption{Few-shot Experimental results}
\begin{tabular}{lc}
   \toprule
    Method & Accuracy(\%) \\
   \midrule
    LLaMA-2(1-shot) & 50.9 \\
   LLaMA-2(2-shot) & 49.4 \\
   LLaMA-2(4-shot) & 49.2 \\
   LLaMA-2-chat(1-shot) & 52.9 \\
    LLaMA-2-chat(2-shot) & 53.9 \\
    LLaMA-2-chat(4-shot) & 51.0 \\
   ChatGPT(1-shot) & 74.2 \\
   ChatGPT(2-shot) & 75.7 \\
   ChatGPT(4-shot) & 75.2 \\
   Codex(4-shot)\cite{datar} & 72.6 \\
   \midrule
   Random Guessing &  50.0 \\
   Human & 92.1 \\
   \bottomrule
\end{tabular}
\label{fewshot-result}
\end{table}


In this paper, we mainly focus on the original capability of LLMs for TFV, directly using them as table-based fact-checkers. Therefore, the methods\cite{binder,datar} using LLMs as additional modules to assist traditional table models are out of the scope of this paper. 

\section{Instruction Tuning of LLMs} \label{sec6}



The LLMs are not trained under the above in-context learning setting, where the focus is on the design of prompts. On the contrary, this section discusses the gain brought by model training and parameter updating (i.e., instruction tuning) of LLMs in TFV under the fine-tuning setting.  
 
\begin{figure}[h]
\centering
\includegraphics[width=1.0\linewidth]{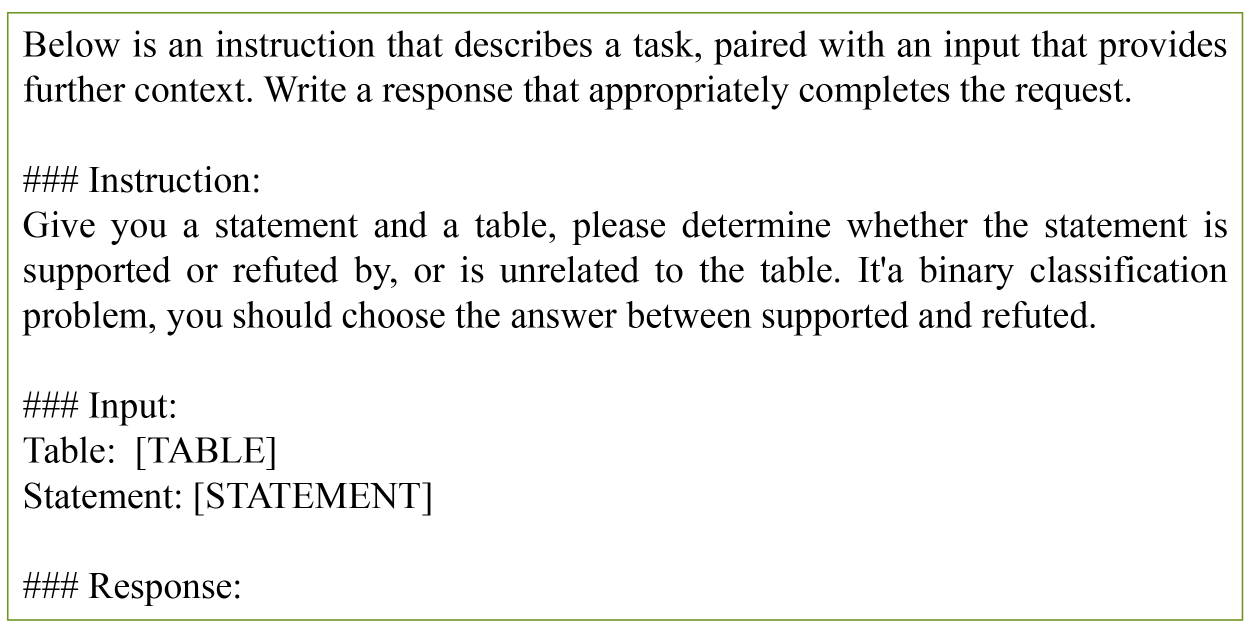}
\caption{Alpaca instruction-following prompt for TFV}
\label{fig:6-1}
\end{figure}

For instruction tuning, we should first construct TFV instruction data, each of which is composed of instruction, input, and output. Followed by Stanford Alpaca\cite{alpaca}, we prepare a TFV instruction and fill the table and statement in the instruction-following format shown in Fig.\ref{fig:6-1}. Then LLMs are fine-tuned with next-word prediction object to predict the verdict words (e.g. supported) in a supervised manner. At the inference stage, we extract the generated content and map it into expected label set. Additionally, considering that tables are long context, we train LLaMA with PEFT method LoRA\cite{lora} instead of full fine-tuning to reduce training cost, which learns pairs of rank-decomposition matrices with other original weights frozen. 

 We choose LLaMA-1 and LLaMA-2 to perform instruction tuning on the train set of TabFact with 92283 samples. The models are fine-tuned on two A100 GPUs. We compare our instruction tuning methods with existing well-trained LLM-free baselines. The results of different fine-tuning methods on TabFact-small-test are shown in Table \ref{finetuned-result}.

\begin{table}[h]
\centering
\caption{Experimental results of Fine-tuning Methods}
\begin{tabular}{llc}
   \toprule
   Type & Method & Accuracy(\%) \\
   \midrule
   \multirow{8}{*}{LLM-free} & Table-BERT\cite{tabfact} & 68.1 \\
   & LPA\cite{tabfact} & 68.9 \\
   & LFC (LPA-based)\cite{LFC}& 74.2 \\
   & ProgVGAT\cite{progvgat} & 76.2 \\
   & BART (large) \cite{tapex} & 82.5 \\
   & Tapas (large)\cite{tapas} & 83.9 \\
   & Tapex (large)\cite{tapex} & 85.9 \\
   & \textbf{PASTA}\cite{pasta} & \textbf{90.6}\\
   \midrule
   \multirow{2}{*}{LLM-based}  & LLaMA-1 & 79.5 \\
   & LLaMA-2 & 82.3 \\
   \midrule
   \textbf{Human} & - & \textbf{92.1} \\
   \bottomrule
\end{tabular}
\label{finetuned-result}
\end{table}

Results show that fine-tuned LLaMA models gain a significant improvement. They outperform certain program-based methods and encoder-only general PLMs but fall short compared to small-scale table-based PLMs. This discrepancy may be attributed to LLaMA models not having exposure to billions of tabular data. 

Moreover, we observe that LLaMA-2 is slightly inferior to BART, a general encoder-decoder PLM with only 0.44 billion parameters. It can be interpreted from two aspects. On the one hand, the LoRA strategy we use only allows a few model parameters to update during fine-tuning, which may prevent to realize the full potential of LLMs. On the other hand, some small language models may be effective enough to perform on individual dataset, while LLMs will exploit their dominance under cross-domain and out-of-domain settings.

\section{Analysis and Discussion} 

\subsection{Case Study}


\begin{figure}[h]
\centering
\includegraphics[width=1.0\linewidth]{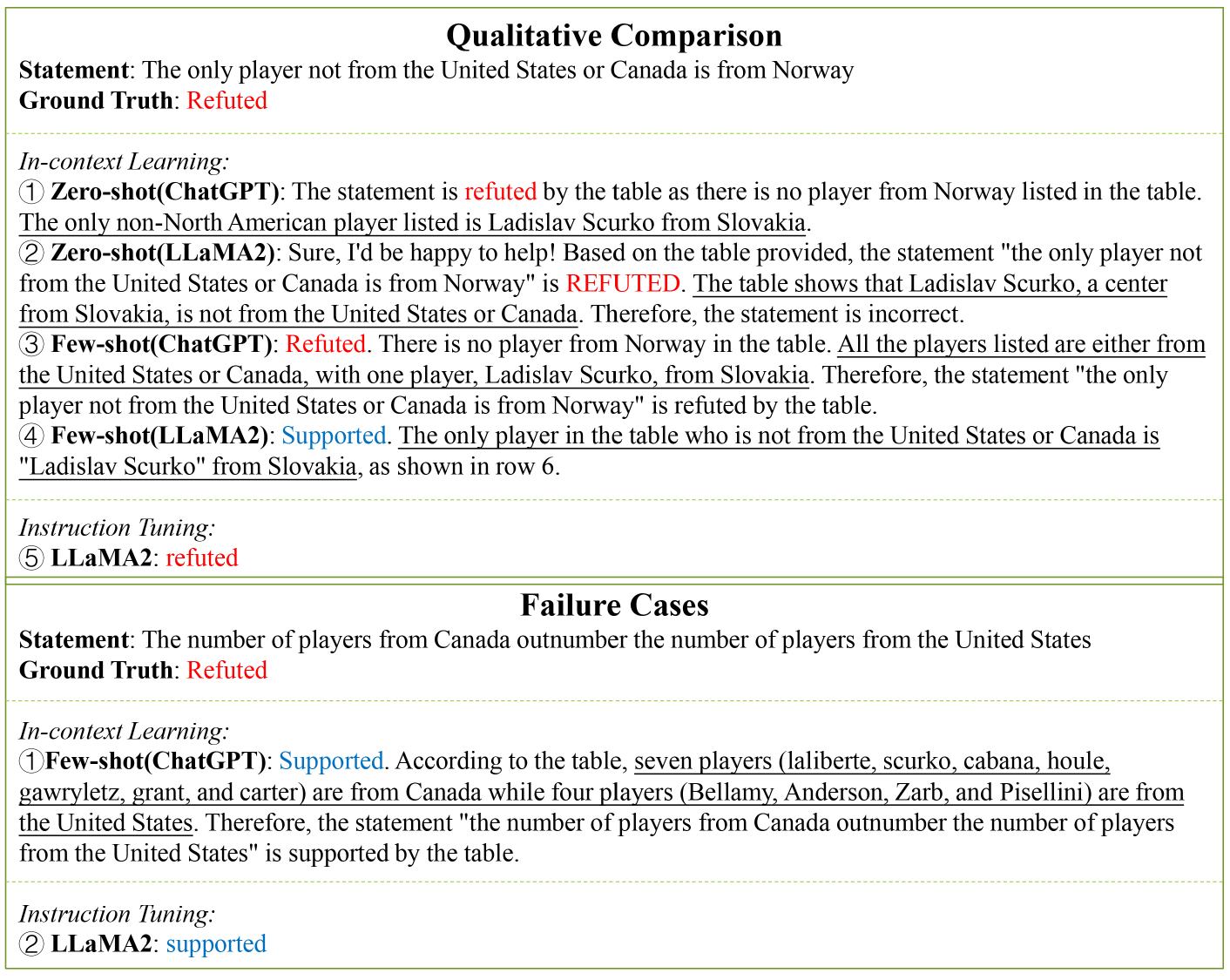}
\caption{Some real responses for two cases towards the table in Fig.\ref{fig:1-1}. We show the responses of best-perform prompt (with two in-context examples) and LLaMA-2 version (LLaMA-2-chat) under the few-shot setting.  } 
\label{fig:7-1}
\end{figure}


Fig.\ref{fig:7-1} upper part reports the qualitative comparison of models under different settings to the same question, which shows that: 1) From zero-shot to few-shot, and then to instruction-tuning settings, the target prediction label (e.g., refuted) gradually move forward in the entire output (from the middle to the beginning), and the generated content of LLaMA2 is more concise and specific, making the answers easier to extract. 2) The response of LLaMA2 under the few-shot setting generates correct explanations but a wrong label, which shows the weak reasoning ability on TFV of LLaMA2 without tuning. 3) Although the accuracy of LLaMA2 is clearly improved after instruction tuning, and the responses are more in line with the format set by TFV task, relevant explanations are no longer generated after instruction tuning due to the lack of explanations annotation during model training. We also showcase the failure examples in Fig.\ref{fig:7-1} lower part. There are actually five players from each of the two countries, however, both the best in-context learning and instruction-tuning results fail to generate the correct labels. This shows that LLMs may suffer severely from the hallucinations for complex (e.g., counting and reasoning) questions. 


\subsection{Discussion}

From the above, we notice that even fine-tuned LLMs still lag behind with some LLM-free methods toward classification accuracy on the TFV task. We would like to discuss some possible research directions in future work to improve the performance on TFV by means of advanced LLMs. 



%
\noindent\textbf{Handling Long Input}.  
The improvement on TFV for LLaMA-2 compared to LLaMA-1 can be primarily attributed to the extension of context windows. It's a universal problem for all tabular tasks to handle excessively long tables. One direct approach is using LLMs with robust extrapolation ability, but it appears to overlook the characteristic of table structure. Besides, a more efficient way is to decompose tables into sub-tables related to their corresponding statements. Datar\cite{datar} first proves the efficacy of LLM-based table decomposers, and more decomposing strategies need to be further explored.  

\noindent\textbf{Specifying Inference Procedures}. Followed by the thought of Chain-of-thought (CoT)\cite{cot}, developing detailed inference procedures in stages, rather than focusing solely on labeling individual words, has potential to enhance performance. For instance, it's effective for LLMs to decompose complex sentences into simpler ones\cite{datar}, which benefits both the accuracy and interpretability for TFV task. We have conducted some preliminary works about LLMs's ability on TFV, but we suffer from the lack of annotated CoT datasets about tables. Further explorations on CoT of TFV are reserved for the follow-up studies.

\noindent\textbf{Developing table-based LLMs}. Inspired by the outstanding performance of small-scale table-based PLMs, it's promising to fine-tune LLMs with intermediate tables reasoning tasks. To develop table-based LLMs, it's necessary to construct instruction datasets of various table types or formats, such as hierarchy tables or HTML-format tables. We infer that TFV under cross-domain and out-of-domain settings will benefit more from table-based LLMs.

\section{Conclusion}

In this paper, we motivate LLMs to perform the table-based fact verification task directly under in-context learning and instruction-finetuning settings. 
Experimental results show that LLMs can be qualified for table-based fact checkers. Larger and intelligent LLMs like ChatGPT can achieve an acceptable result on TFV with elaborate prompts under zero-shot and few-shot settings, while LLMs with small parameters like LLaMA can't. However, the result of LLaMA can be promoted significantly by instruction fine-tuning, though they still lag behind with the most advanced task-specific small-scale models on TFV. 
In addition, 
we also conclude three impressive research directions for the follow-ups.  

\bibliographystyle{IEEEtran}
\bibliography{IEEEabrv, my.bib}


\end{document}